# Optimizing Domain-Specific Image Retrieval: A Benchmark of FAISS and Annoy with Fine-Tuned Features

MD Shaikh Rahman, Syed Maudud E Rabbi, Muhammad Mahbubur Rashid

*Abstract*— Approximate Nearest Neighbor search is one of the keys to high-scale data retrieval performance in many applications. The work is a bridge between feature extraction and ANN indexing through fine-tuning a ResNet50 model with various ANN methods: FAISS and Annoy. We evaluate the systems with respect to indexing time, memory usage, query time, precision, recall, F1-score, and Recall@5 on a custom image dataset. FAISS's Product Quantization can achieve a precision of 98.40% with low memory usage at 0.24 MB index size, and Annoy is the fastest, with average query times of 0.00015 seconds, at a slight cost to accuracy. These results reveal trade-offs among speed, accuracy, and memory efficiency and offer actionable insights into the optimization of feature-based image retrieval systems. This study will serve as a blueprint for constructing actual retrieval pipelines and be built on fine-tuned deep learning networks and associated ANN methods.

## I. INTRODUCTION

Efficient and accurate data retrieval in large-scale datasets is becoming a cornerstone of modern technology, underpinning applications across a wide spectrum of domains [1]. From recommending personalized content in e-commerce [2] to retrieving similar medical images for diagnosis [3], the ability to rapidly and reliably retrieve relevant information from vast repositories is critical. This need has become increasingly urgent with the ever-exploding volume of data, with digital datasets growing at a pace never seen before. Traditional exact search methods, while accurate, often cannot scale to the huge volumes of data involved, especially when data resides in high-dimensional spaces. It is here that Artificial Neural Network (ANN) search techniques have emerged as indispensable tools that allow for scalable and efficient data retrieval [4].

In general, how ANN works is to find, given a query point, the data points in a dataset that are approximately most similar under some defined similarity metric. It is this approximation that allows for significant speedups compared to exact methods, usually at minimal cost in terms of accuracy [5]. As a result, ANN algorithms have seen wide applicability in tasks involving high-dimensional embeddings, such as image and video search, natural language processing, and recommendation systems [6]. Among the many frameworks developed for ANN, FAISS (Facebook AI Similarity Search) [7] and Annoy (Approximate Nearest Neighbors Oh Yeah) [8] remain two of the most popular and widely adopted solutions. Their architecture represents a good balance between retrieval accuracy and computational and memory efficiency necessary to meet real-world limitations linked to latency, scalability, and hardware.

FAISS boasts of its flexibility and the extensive support of advanced indexing methods. It provides a rich set of indexing techniques, including Flat Index, Product Quantization (PQ), and Inverted File Systems (IVF), each targeted at different trade-offs between precision and efficiency [7]. FAISS is particularly favored in scenarios where the retrieval process must scale to billions of data points while maintaining high accuracy. On the other hand, Annoy is highly optimized for low-memory usage and very fast queries. Hence, it is preferred for low-latency applications such as real-time recommendations and search functions.

While both have enjoyed widespread adoption due to robust performance, most related works up until now focused exclusively on the indexing mechanisms and retrieval performance of either FAISS or Annoy. These usually build on either pre-computed or idealized feature embeddings that are independent of the ANN framework to be evaluated [9]. In practical systems, however, feature embeddings are neither static nor universally optimal. Instead, they typically originate from deep learning models, which have to be fine-tuned for the particular dataset and retrieval task [10]. The quality of the features thus produced is one of the most important factors that determine the overall performance of the retrieval system, since it affects directly both the precision and the efficiency of ANN search. However, such a link between feature extraction and ANN indexing has not been considered much in the literature yet. Besides, real-world retrieval systems have constraints well beyond indexing speed and accuracy alone. The computational cost of feature extraction, the memory footprint of the indexing structures, and query time are similarly crucial metrics that impact real-world deployment. Any evaluation of ANN methods in isolation from the feature extraction pipeline is inherently incomplete, as it fails to capture the end-to-end performance characteristics that define actual use cases.

These lacunae are addressed herein in an omnibus manner with FAISS and Annoy in a unified pipeline involving both feature extraction and ANN indexing. In particular, we will finetune a deep learning model on a custom dataset of images to produce feature embeddings that are optimized for the task at hand; these will then be indexed through various configurations of both FAISS and Annoy to afford us the opportunity for systematic benchmarking of the performance of each system. This work provides a clear insight into the practical trade-offs of state-of-the-art feature extraction techniques combined with ANN indexing methods and hence goes beyond the scope of a traditional benchmarking study. By evaluating these methods in an end-to-end manner, we aim to uncover the nuanced interplay between feature quality, indexing efficiency, and retrieval accuracy. Our findings are not only relevant to researchers studying ANN methods but also to practitioners building real-world retrieval systems, where constraints on latency, memory, and computational resources must be carefully managed.

*A. Objectives*

Our main goal is to improve how well approximate nearest neighbor search methods work in a complete system. This system handles everything from feature extraction to indexing. Usually, when we evaluate these search methods, we don't think about how much the quality of features we're using affects them. We tend to assume that if there were better features, the ANN methods would perform much better. We also overlook how practical problems with the features and the ANN methods themselves limit what we can do in real-world applications.

- To assess FAISS and Annoy: we decided to compare them across several indexing methods. These methods are Flat Index, Product Quantization (PQ), Inverted File Systems (IVF), and some combinations of these methods. We looked at performance metrics to do this. These include indexing time, memory usage, query time, precision, recall, and the F1-score. Overall, we wanted a thorough comparison of these two libraries across lots of things.
- Combining Feature Extraction with Searching Methods That Use Artificial Neural Networks: To improve a deep learning model on a specific set of images, we make sure the model creates image features that are good for searching. We want to see how these features affect the image retrieval process when we test different approximation methods that use artificial neural networks.
- Analyzing System Performance: From Start to Finish: To compare FAISS and Annoy not just as indexing tools, but as important parts of a complete retrieval system, we need to look closely at how they work in our system for searching and identifying characters in ancient handwritten texts. This means looking at how well they match our needs during both the indexing and retrieval steps under different conditions.
- Practical Trade-Offs: We look at how retrieval speed, indexing efficiency, and accuracy relate to each other in our experiments. This lets us see how different setups perform and find the best choices for real-world situations.

Setting up design rules for retrieval systems: Make sure the advice we give is clear so people can understand it and use it easily. Give good suggestions about choosing and setting up ANN indexing methods with feature extraction models for certain situations.

This study aims to connect the performance goals that researchers set for ANNs with what real retrieval systems need to work well in practice. We hope this connection will help both researchers and people in industry who use ANNs in real systems.

*B. Motivation*

The rapid growth of image-centric applications has fueled the demand for efficient and accurate image retrieval systems. At main challenge of these systems is to quickly identify similar items from vast datasets, a task that has become increasingly crucial with the exponential growth of data. Approximate Nearest Neighbor (ANN) search methods, such as those implemented in FAISS and Annoy, have emerged as powerful tools to address this challenge, offering scalable solutions with competitive accuracy and speed.

However, real-world image retrieval systems do not operate in isolation. They rely on a combination of robust feature extraction techniques and indexing algorithms. Despite the wealth of research on ANN methods, the interplay between feature extraction models and indexing methods remains underexplored. Most existing studies focus either on indexing or feature extraction in isolation, failing to consider how the quality of features and indexing performance collectively impact the end-to-end retrieval pipeline.

Moreover, the diverse use cases for image retrieval demand a nuanced understanding of trade-offs between indexing speed, memory efficiency, and retrieval accuracy. For instance, an e-commerce platform prioritizing high retrieval precision for user satisfaction may require different configurations compared to a surveillance system focusing on real-time responses. Existing literature often evaluates indexing methods under idealized conditions, leaving a gap in practical guidance for deploying these methods in real-world scenarios with specific constraints. This study is motivated by the need to fill this gap by providing a holistic evaluation of FAISS and Annoy within the context of a complete retrieval system. By integrating a fine-tuned deep learning model for feature extraction with ANN indexing methods, we aim to offer actionable insights into optimizing retrieval performance for practical applications. This research aspires to bridge the divide between theoretical benchmarks and operational needs, contributing to the development of more effective and efficient retrieval systems across diverse domains.

II. BACKGROUND AND RELATED WORK

*A. Approximate nearest neighbor*

Approximate nearest neighbor search is among the very few foundations of modern data-driven applications that enable efficiency in similarity search over high-dimensional vector space. It has been in wide usage for image retrieval, recommendation systems, and information retrieval, where, very often, the task reduces to finding the nearest neighbors of a query point among a massive set of data points [11], [12]. This traditional exact search method has problems due to computational inefficiency when it comes to handling large, high-dimensional datasets [13]. ANN indexing relaxes some of the exactness for speed and memory efficiency. Most ANN methods rely on two methodologies: vector compression and specialized indexing structures. Vector compression reduces the memory footprint and computational cost by approximating vector representations through quantization or binary encoding [14]. Specialized indexing techniques, including but not limited to inverted file systems, graph-based structures, and tree-based partitioning, confine the search space to subsets of vectors, further optimizing query performance [15]. Inverted file indexes, such as IVF and its extensions, group vectors into clusters based on proximity to centroids. Query searches are confined to a few nearest clusters, balancing accuracy and efficiency [16]. Graph-based methods, including HNSW, use hierarchical proximity graphs to traverse layers of increasingly dense vector connections for high recall rates with reduced query times [17]. Other techniques, such as ParlayANN, optimize graph-based methods for parallel processing, scaling to billion-scale datasets with deterministic outputs [18].

Despite these advantages, the choice of ANN technique is highly dependent on the context. For example, while graph-based approaches give superior recall and throughput tradeoffs, their construction can be computationally expensive[19]. Adaptive termination methods, recently proposed, dynamically adjust query depths toward an optimal tradeoff between latency and accuracy for a given dataset and use case [20].

These developments render ANN indexing very important; it bridges the gap between computational efficiency and realistic demands of large-scale applications. The present study has been performed on top of those pioneering methods in order to benchmark state-of-the-art ANN libraries, such as FAISS and Annoy, in front of the Fashion dataset for their indexing capabilities and how suitable they are for real-world scenarios.

*B. FAISS Indexing and Applications*

The Faiss library from Meta has emerged recently as one of the most flexible and efficient similarity search tools for high-dimensional spaces. It mainly strikes a good balance between indexing speed, query efficiency, and memory use, making it very versatile across various applications [7] [21]. Faiss supports several indexing methods, including Flat Index, Product Quantization (PQ), and Hierarchical Navigable Small Worlds (HNSW), hence making it suitable for large-scale data indexing and retrieval tasks with computational efficiency [21] [12].

One of the most salient features of the library is its support for GPU acceleration that gives a tremendous boost in the speed of high-dimensional vector computations. This is very useful for real-time applications that have to deal with enormous volumes of data. In fact, Faiss has been shown to scale well up to billion-sized datasets by using optimizations such as non-exhaustive search and vector compression techniques [21]

Besides technical strengths, there has been wide adoption of Faiss into many domains such as image and text similarity search, recommendation systems, and content moderation [22]. Its application was further broadened by the flexibility that it had to support several distance metrics such as the L2 distance and inner product similarity. Also, recent improvements have focused on integrating it into modern machine learning workflows so it would guarantee seamless interoperability with deep learning frameworks [23].

While powerful, optimization for specific use cases is still very challenging in Faiss. For example, fine tuning the trade-off between speed and accuracy requires hand-picking an indexing strategy along with its hyperparameters. Nevertheless, due to its modular architecture and thorough documentation, it has found applications in both academic research and industrial applications.

*C. ANNOY Indexing and Applications*

Annoy is a tree-based indexing framework for efficient ANN in high-dimensional spaces. It is lightweight and, because of the very low memory overhead when dealing with enormous amounts of data, it can be applied to big applications. It divides the data by means of hierarchical random projection trees to achieve the optimal compromise between retrieval accuracy and query speed [24].

Annoys indexing mechanism is based on recursively splitting the data space using hyperplanes determined by random projection vectors. Each split reduces the search space, enabling logarithmic query time complexity in terms of the number of data points. Multiple trees are constructed independently to increase the likelihood of finding the true nearest neighbors while maintaining computational efficiency [25]. Unlike KD-trees, which suffer from the curse of dimensionality, Annoy does rather well for high-dimensional data; thus, it is more useful for many application fields such as recommendation systems and clustering [26]. It comes along with a very important benefit which is a really memory-friendly implementation; the data points in the index are stored as binary files. This keeps its memory footprint at a minimum while creating and querying the index and thus makes it quite deployable on resource-constrained systems. For instance, Spotify is one of the larger applications which have used Annoy at scale for powering personalized recommendations in real-time.

Empirical evaluations show that Annoy does really well for tasks where the system has to tradeoff between speed and retrieval accuracy. A user can easily tune the number of trees in the ensemble to balance the framework between either precision or computational efficiency. Various studies have suggested that as the number of trees goes up, so does the accuracy, but with added query latency[24]. However, in high-throughput systems, the performance of Annoy remains competitive, outperforming graph-based methods with constant query times in certain benchmarks [27]. Annoy's adaptability extends to various real-world datasets, where its hierarchical tree structure ensures consistent performance irrespective of data distribution. Its performance has been compared to other ANN methods, revealing strengths in query speed for lower recall thresholds and its limitations in handling extremely high-recall tasks compared to graph-based approaches like HNSW.

These characteristics highlight Annoy's suitability for applications prioritizing speed and scalability, such as content recommendation and image retrieval systems. Annoy's lightweight, efficient, and scalable indexing methodology offers a compelling solution for approximate nearest neighbor search. By leveraging hierarchical partitioning and compact data storage, it provides a versatile tool for handling high-dimensional data at scale, with proven efficacy in real-world applications like music recommendation and search systems.

III. METHODOLOGY

*A. Dataset*

In this work, we used Fashion Product Images Dataset, download from Kaggle [28]. The dataset is quite extensive when it comes to fashion-related product images, and thus serves as a good basis for respective image classification and retrieval tasks. The original dataset contained 44,446 images across several categories, and extensive preprocessing was required for these data to suit the purposes of this study and mitigate challenges such as category imbalance and under-representation in some groups. These refinements were needed to ensure that the training, validation, and testing were meaningful and would not result in overfitting or underperformance due to sparse samples in certain subcategories.

To balance the dataset, we reduced original 44 subcategories to 32 subcategories. The approach consisted of merging categories that share similar visual properties. Examples of such categories are Scarves and Mufflers, which have been combined because of shared visual properties, while Sports Equipment and Sports Accessories were put under one category labeled as Sports Equipment. Similarly, Eyes and Eyewear were combined into one category- Eyewear, while Hair and Headwear fell under the category of Headwear. Other categories, like Skin and Skincare, were also merged into one- Skin Care. We also split some into distinctive groups, and removing those categories that have little data to provide greater granularity. For example, *Innerwear* was divided into three distinct subcategories: Innerwear, representing standalone products, and Innerwear_Men and Innerwear_Women, which included images of male and female models wearing the items, respectively. Similarly, *Top Wear* was distributed into three subcategories: T-Shirts, Shirts, and Topwear, allowing for a more detailed analysis of these groups. Some categories, such as *Free Gifts* and *Home Furnishing*, were excluded altogether due to limited samples, while *Apparel Sets* were split into more specific accessory categories like Belts, Tie.

The dataset presented challenges similar to real-world scenarios, such as variations in image quality, resolution, and background. Additionally, overlapping visual features between certain categories, such as Shirts and Topwear or Scarves_Mufflers, posed significant hurdles for classification. Similarly, subtle distinctions within categories, such as Skin Care and Headwear, required a robust model to capture fine-grained details effectively. By restructuring and redistributing the dataset, we manage to mitigate these challenges. This refined dataset forms a robust foundation for evaluating the fine-tuned model and assessing the performance of various approximate nearest neighbor methods.

*B. Model Fine-Tuning*

The ResNet deep architecture is supplemented with residual connections to tackle the vanishing gradient problem in deep neural networks. This enables ResNet to learn robust and discriminative features from complex datasets such as the Fashion dataset more effectively, directly improving the quality of the embedding generated for ANN indexing. Net deserves a specialized focus, as the model showed superior performance over traditional models in various tasks such as image classification, object detection and embedding generation requiring higher accuracy and more efficient feature representation. This enables it to generalize well across tasks.

**Dataset Preparation and Split**

A 90:10:10 train/validation/test split was employed on the 32 subcategories of this dataset. The training set contained most of the data, which allowed sufficient representation of the categories for the model to learn. The validation set monitored model performance during training, and the test set served as the performance benchmark for the final model. This division offered a balanced approach towards model optimization as well as assessment while ensuring that the risk of overfitting is minimal.

**Training Setup and Configuration**

The training was performed for several epochs, defined as a full pass through the training dataset. The model fine-tuned itself in each epoch, updating its parameters to reduce the loss between the predicted output of the model and the actual output used in training, defined by the selected loss function. The Adam optimizer was utilized for optimization, which is a widely used adaptive method for stochastic optimization.

Adam combines the benefits of momentum and adaptive learning rates [29]. At its core, it computes moving averages of the gradients ($m_t$) and the squared gradients ($v_t$) to adjust the learning rate dynamically for each parameter. The update rule for a parameter $\theta\_t$ at step $t$ can be expressed as:

$$m_t = \beta_1 m_{t-1} + (1 - \beta_1) g_t \tag{1}$$

$$v_t = \beta_2 v_{t-1} + (1 - \beta_2) g_t^2 \tag{2}$$

$$\hat{m}_t = \frac{m_t}{1 - \beta_1^t}, \quad \hat{v} = \frac{v_t}{1 - \beta_2^t} \tag{3}$$

$$\theta_{t-1} = \theta_t - \alpha \frac{\hat{m}_t}{\sqrt{\hat{v}_t + \epsilon}} \tag{4}$$

Here, $g_t$ is the gradient of the loss function at time step $t$, $\beta_1$ and $\beta_2$ are exponential decay rates for the moment estimates, $\alpha$ is the learning rate, and $\epsilon$ is a small constant for numerical stability.

The learning rate, a critical hyperparameter that governs the step size of weight updates, was initially set to 0.001. To dynamically adjust this learning rate during training, the ReduceLROnPlateau scheduler was employed. This scheduler monitors the validation loss, reducing the learning rate by a factor (in this case, 0.1) when the validation loss stagnates for a certain number of epochs, as defined by the patience parameter. The reduction in the learning rate can be expressed as:

$$\alpha_{new} = \alpha \times factor \quad (5)$$

where $\alpha_{new}$ is the adjusted learning rate and $factor$ is typically a value like 0.1.

The training process also included gradient clipping, where the magnitude of gradients was restricted to a maximum norm of 1.0. This technique helps prevent exploding gradients, which can destabilize the training process, especially in deep networks.

Finally, early stopping served as a regularization mechanism, meaning that the training was stopped when, for three epochs in a row, the validation loss did not improve. This ensured efficient use of resources, preventing overfitting by stopping training at an optimal point in time.

Some performance metrics during training were the following to comprehensively analyze the model's behavior:

**Precision**
Precision calculates a ratio of correctly predicted positive samples to all the samples predicted as positive. This is very useful when there are significant consequences of false positives [30].

$$Precision = \frac{True\ Positives\ (TP)}{True\ Positives\ (TP) + False\ Positives\ (FP)}$$

**Recall**
Recall is a metric that tests the model on correctly identifying real positive samples. It becomes vital when the cost of missing true positives is high [31].

$$Precision = \frac{True\ Positives\ (TP)}{True\ Positives\ (TP) + False\ Negatives\ (FN)}$$

**F1-Score**
The F1-Score is the harmonic mean of precision and recall. It gives one measure which considers both aspects, extremely useful in cases of an imbalanced dataset [32].

$$F1\ Score = 2 \times \frac{Precision \times Recall}{Precision + Recall}$$

**Accuracy**
Accuracy is the overall ratio of properly classified samples, incorporating both true positives and true negatives in relationship to the total number of samples [33].

$$Accuracy = \frac{True\ Positives\ (TP) + True\ Negative\ (TN)}{Total\ Number\ of\ Samples}$$

Taken all together, these metrics have provided different insights into aspects of the model's predictive performance and thus allowed nuanced understanding of strengths and weaknesses while training and validating the models. While accuracy gives the overall measure, precision and recall show how well a model can deal with certain classes, whereas F1-Score captures a trade-off between them.

Validation loss is monitored at the end of each epoch. In case of improvement in the performance of the model, validation loss was getting smaller and smaller, model weights will be saved in best_model.pth.

**Embedding Extraction**
After fine-tuning we used the trained model to extract embeddings from the dataset. These embeddings represent compact, high-dimensional feature vectors capturing the visual characteristics of the images [34]. By passing an image through the fine-tuned ResNet50 and removing the final classification layer, embeddings were obtained from the preceding fully connected layer. These embeddings served as inputs for the approximate nearest neighbor (ANN) methods, facilitating efficient and accurate image retrieval.

This process enabled the model to adapt effectively to the unique characteristics of the fashion dataset, yielding embeddings suitable for downstream retrieval tasks. The integration of these embeddings into ANN methods, such as FAISS and Annoy, enabled a comprehensive evaluation of indexing and querying performance.

*C. ANN Indexing and Query*
In this study, we evaluated Approximate Nearest Neighbor (ANN) methods to understand its effectiveness and efficiency in large-scale image retrieval tasks. ANN indexing and querying techniques were implemented using FAISS (Facebook AI Similarity Search), Annoy (Approximate Nearest Neighbors Oh Yeah), and the HNSW (Hierarchical Navigable Small World) algorithm. The analysis involved comparing their performance across various parameters, such as precision, recall, indexing time, memory usage, and query time.

FAISS is a library developed by Facebook AI. It is highly optimized for similarity searches and clustering of dense vectors. In this work, several indexing methods were implemented based on FAISS which includes Flat-L2, which utilizes the exact Euclidean distance for similarity searches. It also served as the baseline for performance comparison. Another method, Flat-IP, utilizes dot product similarity to rank normalized embeddings. To handle the trade-off between memory usage and retrieval accuracy, Product Quantization, which compresses vectors using quantization techniques, was employed. More advanced indexing methods included Inverted

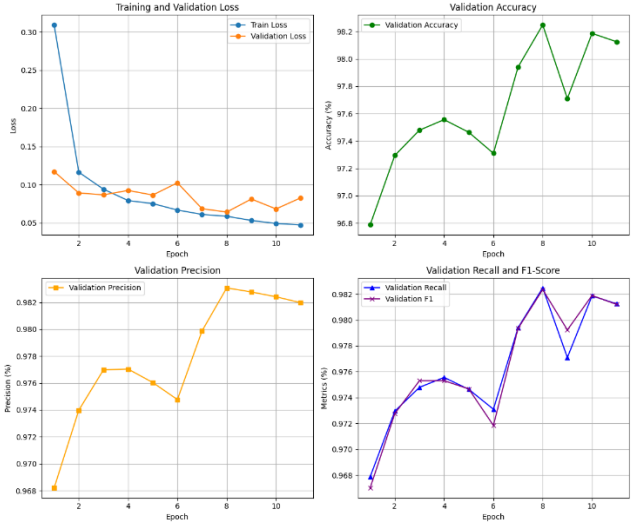

**Fig. 1:** Model performance metrics across 10 epochs, including loss, accuracy, precision, recall, and F1-score for validation data.

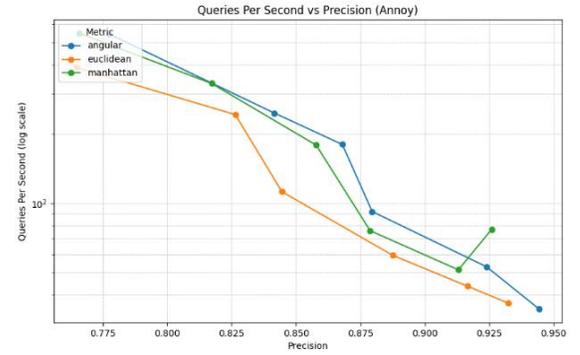

**Fig. 2:** Queries per second versus precision for Annoy with different distance metrics: angular, Euclidean, and Manhattan.

File with PQ (IVF-PQ) for scalability, which combines inverted file indexing with product quantization, and IVF with Scalar Quantization (IVF-SQ), applying scalar quantization to achieve memory efficiency. Locality Sensitive Hashing was also included to offer a hashing-based approach for approximate searches. All these will be tested by extracting the embeddings from a fine-tuned ResNet50 in an effort to understand the strong points and shortcomings of each of these configurations.

Annoy is a Spotify-developed library, and used to implement memory-efficient indices that enable approximate nearest neighbor search. Unlike FAISS, which offers a range of indexing methods, Annoy relies on multiple binary search trees to approximate similarity searches. The distance metrics supported by Annoy, such as angular, Euclidean, and Manhattan, were implemented in this study to compare their effectiveness. The number of trees in the index which is a critical parameter controlling the trade-off between accuracy and memory usage, was set to 10. Annoy's ability to construct compact indices and handle various distance metrics made it an effective alternative to FAISS for certain applications.

Another evaluated algorithm was the Hierarchical Navigable Small World (HNSW), implemented through FAISS. HNSW is a graph-based approach to perform approximate nearest neighbor searches by building a multi-layer graph structure. Each node in the graph corresponds to a data point, with edges representing proximity between points. Key parameters set for HNSW included graph connectivity (M) set to 32, while the recall quality parameter efConstruction was set to 40. HNSW showed a good balance of high recall with efficient memory usage along with quick query times.

*D. Evaluation Protocol*

To ensure a fair and consistent evaluation of the ANN indexing methods, a carefully designed protocol was implemented. The evaluation process began with the selection of 1,000 query images sampled randomly from the test set of the fine-tuned ResNet50 model. By using a fixed subset of query images, the evaluation avoided variability that might arise from different input distributions, ensuring direct comparability of the results across different indexing methods.

The embeddings for all images in the test dataset were extracted using the fine-tuned ResNet50 model. These embeddings served as inputs to the respective ANN indexing methods. Each method was evaluated under identical conditions, including the number of neighbors retrieved for each query (top 6, excluding the query itself). Key performance metrics such as indexing time, query time, memory usage, index size, precision, recall, F1-score, and Recall@5 were recorded for each method. Query performance metrics were computed based on the predictions generated by the top retrieved neighbors for each query.

Although the experiments were executed on a system equipped with an NVIDIA GeForce RTX 3050 GPU, the computations for ANN indexing and querying were carried out on the CPU to ensure compatibility across all methods. This decision ensured that the same hardware resources were utilized for both FAISS and Annoy, as Annoy does not support GPU acceleration. The operating environment are- Python 3.10, PyTorch 2.0 for feature extraction, FAISS 1.7.3 for ANN indexing, and Annoy 1.17.1 for tree-based similarity searches. The indexing and query experiments were conducted on a system running Windows 11, with 16 GB of RAM and an Intel Core i7 processor, ensuring ample computational resources for the experiments.

IV. RESULTS

*A. Model Fine-Tuning Results*

This section shows the training and validation metrics obtained during the fine-tuning process of the ResNet model. It highlights the model's performance and improvements over successive epochs in terms of loss, accuracy, precision, recall, and F1-score.

From Figure 1, we observe the trends in training and validation losses over the 11 epochs. The training loss decreases steadily. Starting at 0.309 in the first epoch and reaching 0.047 by the final epoch. This consistent reduction indicates effective learning by the model. Similarly, the validation loss shows improvement, reducing from 0.117 to 0.083, with minor fluctuations around epochs 4 and 9, possibly due to increased complexity in feature learning. The validation accuracy, as shown in Figure 1, increases steadily, surpassing 98% in the final epochs. This improvement is supported by corresponding increases in validation precision, recall, and F1-score. For example, the precision metric improves from 0.968 in the first epoch to 0.982 in the eighth epoch, with a slight stabilization afterward. The recall and F1-score exhibit similar trends, suggesting balanced improvements in both the model's sensitivity and overall prediction quality.

TABLE 1:
Training and Validation metrics for each epoch

| Epoch | Train Loss | Validation Loss | Validation Accuracy | Validation Precision | Validation Recall | Validation F1 | Epoch Time (s) |
|---|---|---|---|---|---|---|---|
| 1 | 0.309395 | 0.117163 | 96.78709 | 0.968202 | 0.967871 | 0.967016 | 3882.722 |
| 2 | 0.116315 | 0.089146 | 97.29439 | 0.973977 | 0.972944 | 0.972756 | 3982.917 |
| 3 | 0.094103 | 0.086676 | 97.47886 | 0.976979 | 0.974789 | 0.9753 | 4119.838 |
| 4 | 0.079176 | 0.092426 | 97.55573 | 0.977037 | 0.975557 | 0.975302 | 4116.876 |
| 5 | 0.07519 | 0.086424 | 97.46349 | 0.976038 | 0.974635 | 0.974667 | 3996.772 |
| 6 | 0.066727 | 0.102437 | 97.30976 | 0.974769 | 0.973098 | 0.971843 | 4360.788 |
| 7 | 0.061025 | 0.068443 | 97.94005 | 0.979878 | 0.9794 | 0.979345 | 4196.677 |
| 8 | 0.058538 | 0.064095 | 98.2475 | 0.983071 | 0.982475 | 0.982352 | 4310.607 |
| 9 | 0.05312 | 0.081166 | 97.70945 | 0.982772 | 0.977095 | 0.979225 | 4345.242 |
| 10 | 0.049109 | 0.068189 | 98.18601 | 0.982412 | 0.98186 | 0.981893 | 4843.798 |
| 11 | 0.047275 | 0.082706 | 98.12452 | 0.981974 | 0.981245 | 0.981199 | 5312.244 |

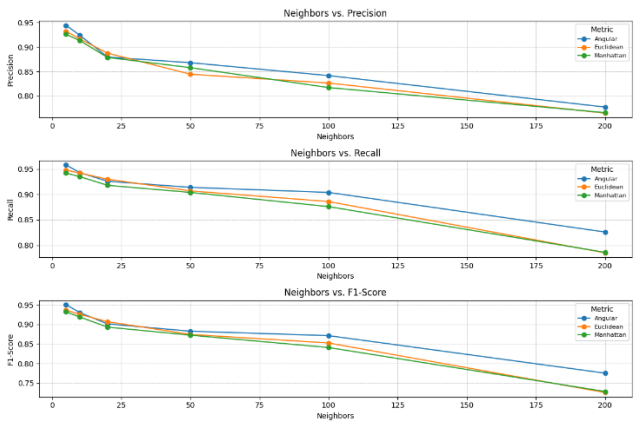

**Fig. 3:** Impact of the number of neighbors on precision, recall, and F1-score for different distance metrics (angular, Euclidean, and Manhattan)

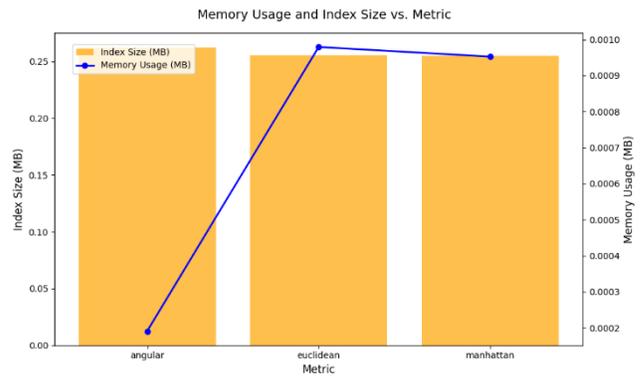

**Fig. 4:** Comparison of index size and memory usage (in MB) for different distance metrics (angular, Euclidean, and Manhattan)

From Table 1, it is evident that both training and validation score improves over time. It confirms the model's ability to generalize well to unseen data. For example, by epoch 8, the validation accuracy peaks at 98.25%, and the corresponding F1-score reaches 98.24%, reflecting the balance in precision and recall.

*B. Ann Indexing Results*

All the ANN indexing methods were evaluated using the angular, Euclidean, and Manhattan metrics provided from the Annoy library. All the evaluations of query speed, precision, recall, memory usage, and index size are shown in the results. Figures 2, 3 and 4 show the different performance based on the metrics above in all these aspects.

Query Speed vs. Precision:
As illustrated in Figure 2, the trade-off between query speed (measured as queries per second) and precision for the three Annoy metrics shows distinct trends. Angular and Manhattan metrics demonstrate better precision at higher query speeds compared to Euclidean. For instance, with 200 neighbors, the Angular metric achieves 0.7774 precision and a query speed of 544.8 QPS, while Manhattan reaches 0.7659 precision at 548.8 QPS. In comparison, Euclidean yields a slightly lower query speed of 388.9 QPS with 0.7649 precision for the same number of neighbors. These trends highlight Angular's and Manhattan's efficiency in balancing speed and accuracy.

Number of Neighbors vs. Accuracy Metrics:
Figure 3 presents the impact of varying the number of neighbors (k) on precision, recall, and F1-score for the three metrics. Angular consistently outperforms Euclidean and Manhattan in accuracy metrics as k increases. For instance, with 50 neighbors, Angular achieves an F1-score of 0.8825, while Euclidean and Manhattan achieve 0.8743 and 0.8724, respectively. Increasing the number of neighbors improves precision but slightly reduces recall and F1-score after a certain threshold. For example, with 200 neighbors, Angular's recall decreases to 0.826, and Manhattan's recall reduces to 0.786, suggesting a balance is required for practical applications.

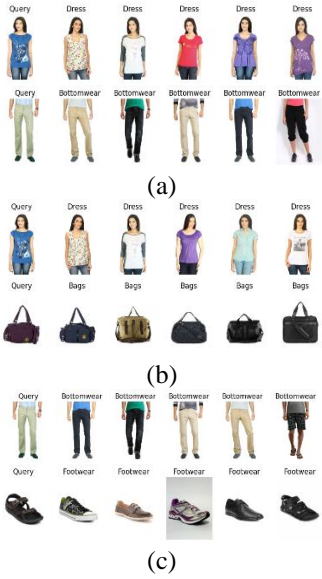

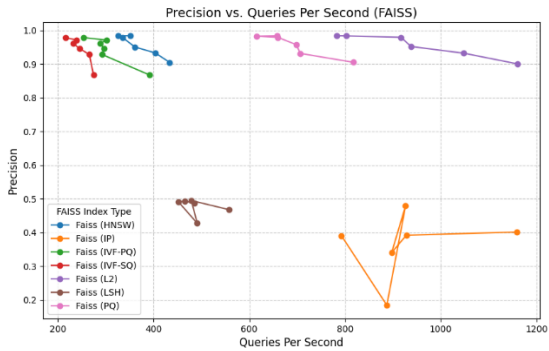

**Fig. 5:** Retrieval results for fashion product queries using different distance metrics: (a) Angular, (b) Euclidean, and (c) Manhattan, demonstrating variations in nearest neighbor selection across categories.

**Fig. 6:** Precision versus queries per second for various FAISS index types (HNSW, IP, IVF-PQ, IVF-SQ, L2, LSH, PQ), showcasing the trade-offs between retrieval speed and accuracy across indexing methods.

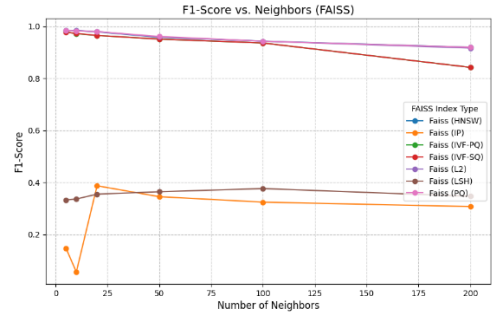

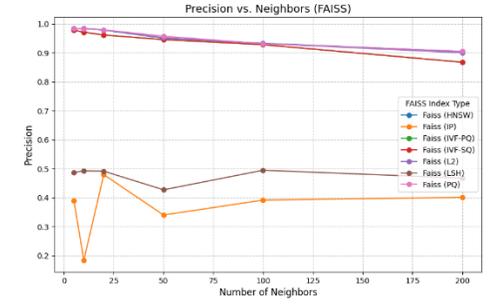

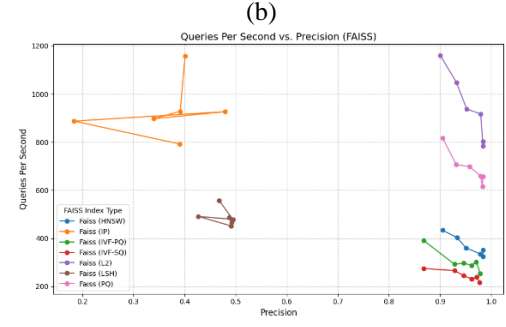

**Fig. 7:** Comparison of FAISS index types: (a) F1-score versus the number of neighbors, (b) precision versus the number of neighbors, and (c) queries per second versus precision, highlighting trade-offs in retrieval accuracy, speed, and neighborhood size for different indexing methods.

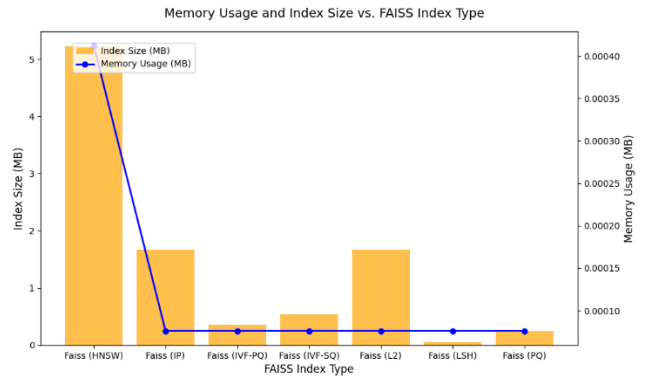

**Fig. 8:** Comparison of index size and memory usage (in MB) for different FAISS index types, highlighting the trade-offs between storage requirements and runtime memory efficiency.

Memory Usage and Index Size:
Figure 4 compares memory usage and index size across the three Annoy metrics. While the index size remains relatively constant across metrics, memory usage varies slightly, with Euclidean requiring the highest memory at 0.00098 MB, followed by Manhattan at 0.00095 MB, and Angular at 0.00019 MB. This difference is attributed to Euclidian's computation-intensive nature for distance calculations. Angular demonstrates a favorable balance, achieving a precision of 0.9443 with minimal memory usage and an index size of only 0.26 MB for 5 neighbors, making it a strong candidate for applications requiring efficiency. Figure 5 shows the retrieval

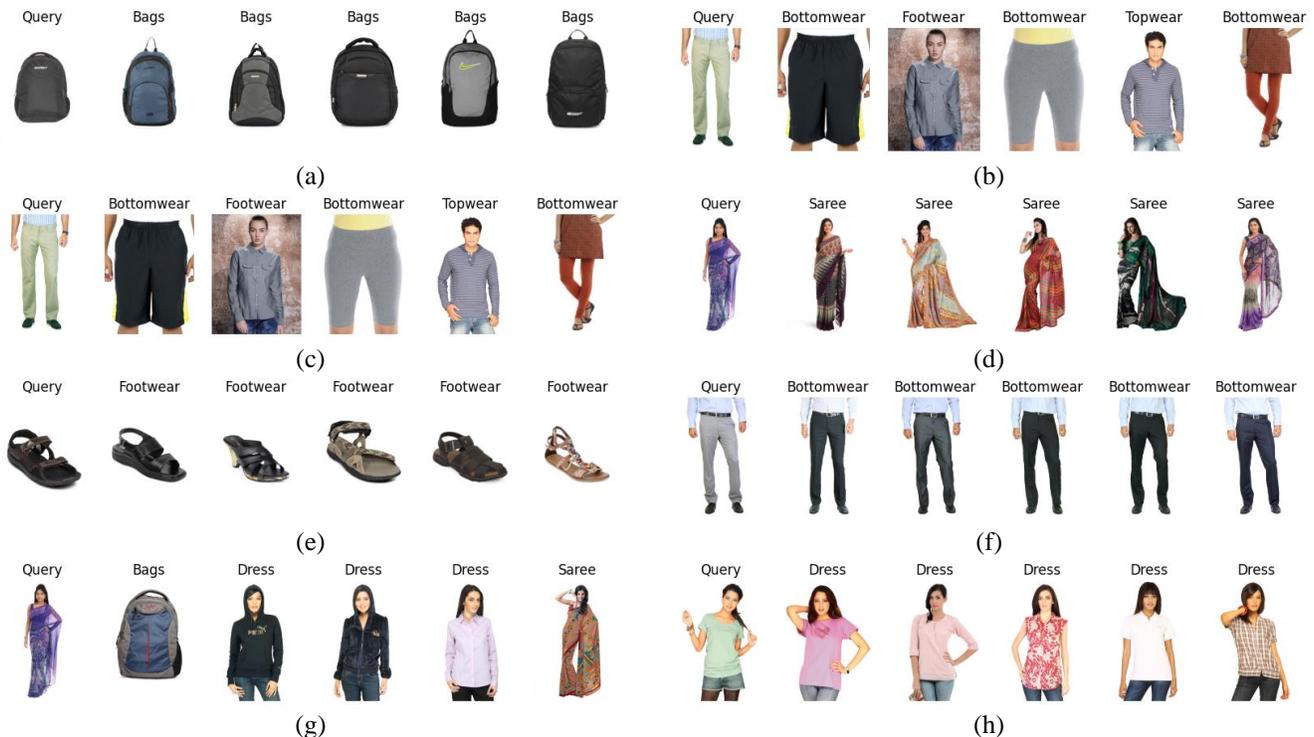

**Fig. 9:** Retrieval results for various indexing methods: (a), (b) Flat Index (L2), (c) Flat Index (Inner Product), (d) Inverted File with IVF, (e) Product Quantization (PQ), (f) IVF with Scalar Quantization, (g) Locality-Sensitive Hashing (LSH), (h) HNSW

results. All metrics of annoy manage to retrieve similar images and access the subcategory successfully.

Angular is shown to be the most effective metric in large-scale data indexing due to its combined effects of high precision, competitive query speed, and reasonable memory use. Euclidean is highly precise in certain cases but it consumes more memory and shows slower query operations. Manhattan gives a good compromise between the query speed and accuracy while offering consistent performance, being somewhat less precise at larger neighborhoods. For example, Manhattan with 100 neighbors reaches a precision of 0.8172 at 331.9 QPS, hence outperforming Euclidean concerning query speed while keeping roughly the same accuracy.
These results, therefore, suggest that the choice of metric should be informed by specific application requirements.

Angular is well suited for applications that require the highest precision and speed, while Manhattan may be better suited for scenarios requiring stable performance across varying neighbor counts.

*C. FAISS Indexing Results*

Figure 6 depicts the precision and speed trade-off for different FAISS indexing methods. Among all the variants, FAISS(L2) has achieved a quite high precision value of about 0.984, while it maintains QPS as high as 803.3 for 5 neighbors. This balance really demonstrates the efficiency in applications that require both accuracy and speed. In contrast, FAISS (IP) gives distinctly lower precision, 0.39, for a comparable QPS of 792.5 at 5 neighbors, reflecting that this algorithm, though very fast, is unsuitable for high-precision requirements. On the other hand, FAISS (PQ) provides competitive precision, 0.984, with a QPS of 657.7, proving its strength for scenarios requiring compact indices with decent accuracy. FAISS HNSW, though yielding a relatively lower QPS 351.1, presents a competitive precision of 0.984, which speaks to its robustness for high-precision tasks at the expense of speed.

Figures 7(a), 7(b), and 7(c) show the impact of increasing the number of neighbors on precision, recall, and F1-score. FAISS (L2) shows a gradual worsening in all metrics as the number increases. For example, from 0.984 for 5 neighbors to 0.9006 for 200 neighbors for precision, while the same trend is followed by the F1-score, which lowers from 0.9838 to 0.9163 within the same range. PQ of FAISS also behaves similarly: it keeps precision at 0.984 with 5 neighbors but drops down to 0.9053 with 200 neighbors. FAISS (HNSW) has a bigger memory footprint, while precision and recall smoothly decrease - the balance between indexing complexity and retrieval performance is reflected.

Figure 8 gives the memory usage and index size of the FAISS methods. FAISS (HNSW) has the largest index size at 5.22 MB, way over that of other methods such as FAISS (PQ) at 0.24 MB. However, in terms of memory usage, FAISS (HNSW) is efficient enough, with 0.00041 MB, to be applicable on systems with tight memory restrictions. FAISS (IVF-PQ) and FAISS (IVF-SQ) manage to strike a balance between compact indices of size 0.35 MB and 0.53 MB, respectively, while being competitive in retrieval performance. These methods are optimal for large-scale datasets where memory efficiency is at the fore.

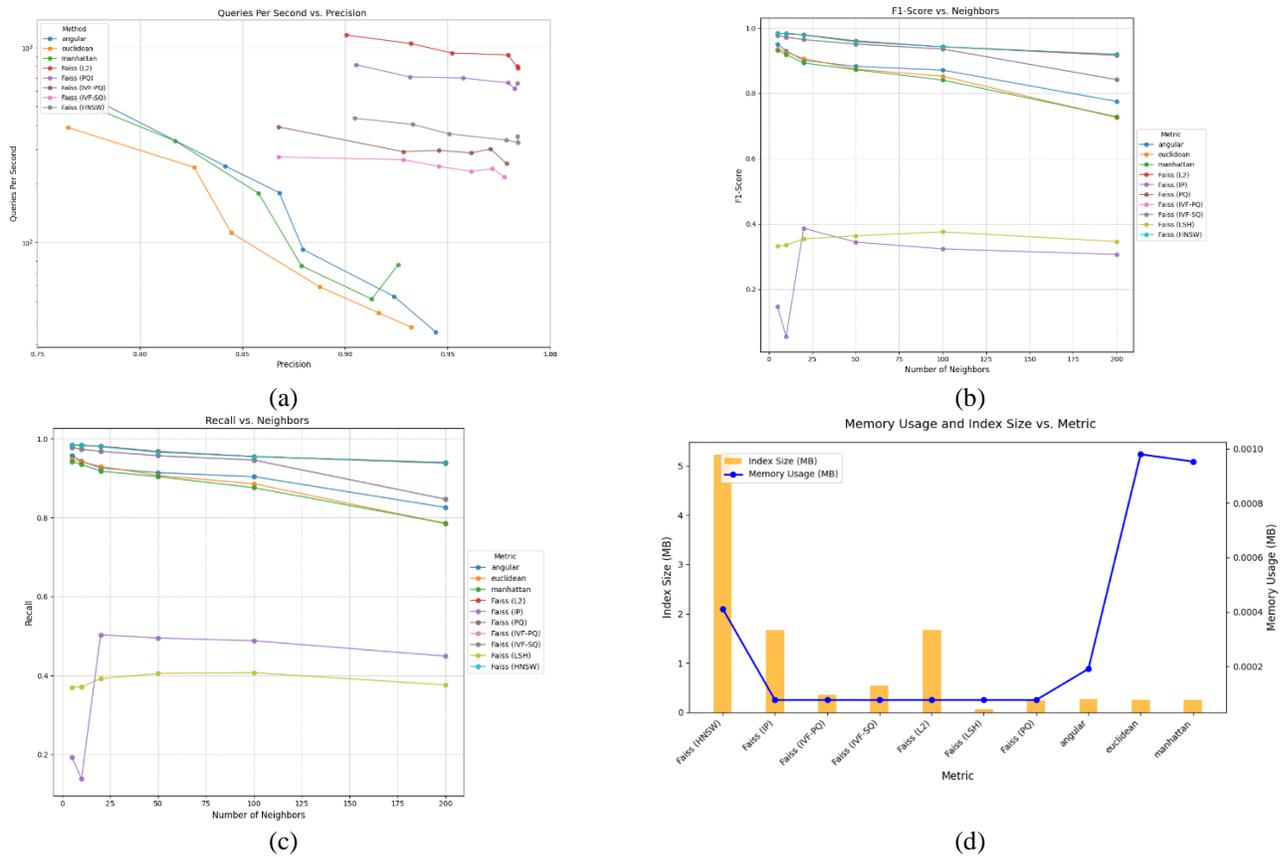

**Fig. 10:** Comparative analysis of performance metrics across indexing methods: (a) Queries Per Second vs. Precision, (b) F1-Score vs. Neighbors, (c) Recall vs. Neighbors, and (d) Memory Usage and Index Size vs. Metric.

From the analysis, the FAISS indexing methods provide a different trade-off for different applications. FAISS (L2) is best for high-precision requirements, especially in finding a balance between precision and speed, reaching 0.984 precision with 1.67 MB index size and QPS of 803.3. FAISS (PQ) has the same precision of 0.984 but with a much smaller index size of 0.24 MB, at the cost of lower QPS, 657.7. FAISS (HNSW) exhibits the largest index size (5.22 MB) but maintains high precision (0.984) and low memory usage, making it suitable for accuracy-prioritized scenarios. FAISS (IVF-PQ) and FAISS (IVF-SQ) offer compact index sizes with decent accuracy, making them ideal for memory-efficient large-scale deployments. The choice of method ultimately depends on the specific trade-offs required by the application, whether prioritizing speed, accuracy, or memory efficiency.

*D. Query Performance*

Query performance was assessed across multiple metrics, focusing on aspects such as memory usage, precision, recall, F1-score, and query speed. Table 2 summarizes the performance metrics for the Annoy and FAISS libraries at k=5, while Figures 10 (a, b, c, d) provide detailed visual comparisons.

Query Speed vs. Precision: Figure 10 (a) highlights the trade-off between query speed and precision. Annoy angular achieved 53 queries per second with a precision of 93.56%. In contrast, FAISS PQ achieved significantly lower queries per second (5.6) but higher precision at 98.4%. This suggests that Annoy provides faster query responses, while FAISS excels in accuracy for certain indexing methods.

Impact of Number of Neighbors: Figures 10 (b) and (c) illustrate the effect of increasing the number of neighbors on recall and F1-score. Annoy angular demonstrates consistent recall improvement with increasing k, achieving 95.4% at k=5. FAISS HNSW achieves the highest
recall of 98.51% at k=5, showing its effectiveness for scenarios prioritizing recall.

Memory Usage and Precision: As depicted in Figure 10 (d), Annoy metrics (angular, Euclidean, and Manhattan) show consistent memory usage, with angular requiring 0.00019 MB, the lowest among the three. FAISS exhibits higher variability in memory usage, with HNSW requiring the most memory at

0.01199 MB. Angular achieved the highest precision among Annoy metrics at 93.56%, compared to 92.94% for Euclidean and 91.95% for Manhattan. For FAISS, precision values were notably lower across most methods, except for PQ and HNSW, which achieved 98.4% and 98.5%, respectively.

**Table 2:**
Performance comparison of various indexing methods at k=5k = 5k=5, evaluating memory usage, precision, recall, F1-score, Recall@5, index size, indexing time, and average query time. The table highlights trade-offs between accuracy, efficiency, and resource utilization across methods like Annoy (Angular, Euclidean, Manhattan) and FAISS (L2, IP, PQ, IVF-PQ, IVF-SQ, LSH, HNSW)

| Metric | Memory Usage (MB) | Precision | Recall | F1-Score | Recall@5 | Index Size (MB) | Indexing Time (ms) | Average Query Time (μs) |
|---|---|---|---|---|---|---|---|---|
| Annoy (Angular) | 0.0001907 | 0.93565575 | 0.954 | 0.94397 | 0.512 | 0.261955261 | 53.52854729 | 16.30067825 |
| Annoy (Euclidean) | 0.0001907 | 0.92941332 | 0.95 | 0.9374 | 0.513 | 0.255157471 | 57.97672272 | 17.3978805 |
| Annoy (Manhattan) | 0.0009984 | 0.91952438 | 0.94 | 0.9276 | 0.51 | 0.254882813 | 72.32570648 | 34.7359180 |
| Faiss (L2) | 0.0000763 | 0.00031565 | 0.98 | 0.984 | 0.983647 | 0.52 | 1.172781 | 1671185.49 |
| Faiss (IP) | 0.0000763 | 0.00033263 | 0.38 | 0.124 | 0.029156 | 0.001 | 1.101971 | 1671185.49 |
| Faiss (PQ) | 0.0000763 | 0.00020050 | 0.98 | 0.984 | 0.983631 | 0.52 | 5.597591 | 240224.838 |
| Faiss (IVF-PQ) | 0.0000763 | 0.00010854 | 0.98 | 0.98 | 0.979813 | 0.518 | 11.414051 | 357730.865 |
| Faiss (IVF-SQ) | 0.0000763 | 0.0000797 | 0.97 | 0.979 | 0.978315 | 0.517 | 3.494501 | 535621.643 |
| Faiss (LSH) | 0.0000763 | 0.00017773 | 0.47 | 0.371 | 0.333183 | 0.312 | 2.0926 | 56218.147 |
| Faiss (HNSW) | 0.01199913 | 0.00011949 | 0.98 | 0.985 | 0.984782 | 0.522 | 316.8408871 | 5222925.18 |

Combined Memory Usage and Index Size: Figure 10 combines memory usage and index size for Annoy and FAISS metrics. On the other hand, the Angular and Euclidean metrics of Annoy have favorable memory/index size trade-offs, with angular showing an optimal balance for large-scale indexing tasks. Among the FAISS methods, HNSW has high requirements in terms of memory and index size, which may show its possible application for tasks that demand higher recall and precision at the cost of greater resource consumption. Annoy angular turns out to be a good competitor in cases when speed is more important but at the cost of some precision, while methods like PQ and HNSW from FAISS will be used when high recall and precision are required. These findings are important for selecting an appropriate method based on specific application requirements.

Indexing Time and Query Time: Table 2 reveals that Annoy angular has the fastest indexing time of 53 ms and an average query time of 16 μs, making it ideal for applications requiring low latency. Conversely, FAISS HNSW, while achieving the highest recall and precision, requires significantly more time, with 316 ms for indexing and an average query time of 5222 μs. This trade-off highlights the suitability of different methods based on application needs.

## V. DISCUSSION

The comparison of the Annoy and FAISS indexing methods in ANN search shows their relative strengths and weaknesses, but each will be more or less suitable depending on the dataset and application characteristics. Fashion dataset is considered in this study-a well-balanced mix of complexity and size, representative for many used to benchmark retrieval tasks.

*A. Comparative Analysis of FAISS and Annoy for Fashion Dataset*

This work presents a novel approach to the evaluation of ANN methods, combining a custom-trained ResNet model with indexing techniques. Unlike traditional studies that often use pre-trained embeddings, this research focuses on the end-to-end pipeline: training a model specifically for the Fashion dataset, generating embeddings, and further assessing indexing and retrieval performance. This allows us to tailor our approach to provide insight specifically relevant to the Fashion dataset, which contains visual classes of moderate complexity-for example, items that are visually similar yet distinct in functionality.

For the Fashion dataset embeddings, Annoy and FAISS were compared on their nearest neighbor retrieval performance. Of the ones tried, Annoy's angular metric was the most resilient, yielding a precision of 93.6% and recall of 95.4% at k=5 with only 16 μs query time. As can be seen in Table 2, this makes angular an excellent candidate for real-time applications were speed and accuracy are important. It showed consistent performance with an increase in the number of neighbors, hence proving its stability and suitability for applications that require uniformity in retrieval metrics. The memory usage for Annoy's angular metric was very low at about 0.0002 MB per query, which further enhances its feasibility for deployment on computationally poor environments such as mobile applications or embedded systems.

In contrast, the methods of FAISS, HNSW, and PQ outperform Annoy by a big margin, having reached up to 98.5% precision and recall for k=5. These gains in accuracy, however, come at the price of significant increase in memory consumption and query time. For instance, HNSW required 0.012 MB per query and showed query times over 5 ms, which is not precisely suitable for applications that require low latency. Meanwhile, the PQ method of FAISS also works well on retrieval metrics, returning precisions and recalls of around 98.4%, while the memory usage and query latency remained higher than that of the Annoy method.

The most salient observation in this work is the trade-off between the number of neighbors and retrieval metrics such as precision, recall, and F1-score. The angular metric of Annoy shows consistent performance for an increase in k, hence showing its robustness. In contrast, the FAISS methods, such as PQ and HNSW, degrade slowly with increasing k, presumably due to their more complex algorithms which scale poorly with the size of the number of neighbors.

Another point is the dataset specificity. Particular challenges exist in Fashion that involve, for instance, similar items, including different shirt and shoe styles, which could only become useful for retrieval if highly precise and of high recall. In the case where the ResNet model trained on this dataset was used for creating embeddings, the feature representations fit with the characteristics inherent in the dataset, to the effect that ANN search indexing could then be implemented in a natural environment of e-commerce fashion-based applications.

Besides, in this context, the metric choice is important. The angular metric proved to be optimal for the Fashion dataset since it balances speed and accuracy pretty well. In such applications as e-commerce search engines or personalized recommendations where responsiveness and user experience matter, angular provides a solution. Conversely, methods in FAISS such as HNSW are applicable to situations in which high precision has a priority over high speed-for example, data analytics pipelines or other decision-making processes where large errors imply serious costs.

The main point underscored here is the suitability of ANN methods for current data and applications. In the Fashion dataset, the overall best choice was Annoy's angular metric due to its competitive retrieval performance with low resource consumption and also showed consistent behavior across varied configurations. However, for use cases with no room for error, where absolute accuracy cannot be sacrificed and computational resources are rich, then FAISS methods like HNSW become good alternatives. This presented work aims to bring a holistic approach towards ANN benchmarking by embedding a custom-trained model using ANN evaluations while keeping the indexing and retrieval methods as close to both data and use-case requirements as possible.

*B. Limitations*

This work has shed sufficient light on the performance of both FAISS and Annoy on the Fashion dataset, though with a few limitations. The analysis was only done on one dataset, which can hardly be enough to project challenges brought about by much larger variability in datasets or even from other modalities such as medical images or videos. As a fact, this limits generalization of results since the method performing best for the Fashion dataset may not yield a similar performance on other datasets.

The evaluation metrics—precision, recall, and F1-score—offer a good understanding of retrieval quality but do not fully account for subjective relevance in practical applications, such as e-commerce search. Additionally, operational factors like scalability for larger datasets and the impact of distributed deployment were not explored

*C. Future Work*

Future work in this domain could focus on expanding the scope of this study explore new avenues for improvement. One key direction is testing FAISS and Annoy on diverse datasets with higher variability, such as medical imaging or 3D point cloud data. This would help determine the robustness and adaptability of these indexing methods.

Additionally, investigating other state-of-the-art ANN libraries, such as ScaNN (Scalable Nearest Neighbors) [35], NGT (Neighborhood Graph and Tree) [36], or HNSWlib [37] could provide a broader comparative analysis. Incorporating hybrid indexing approaches that combine multiple algorithms could also be explored to balance precision, speed, and resource usage.

## VI. CONCLUSION

This work provides an extensive comparison between two of the most popular ANN indexing libraries, FAISS and Annoy, using the Fashion dataset as a benchmark. By combining deep learning-based feature extraction through a ResNet model with ANN indexing techniques, we aimed to provide insight into the practical trade-offs that exist between precision, speed, memory usage, and scalability. The results clearly indicated that both FAISS and Annoy had different strengths that made them more suitable for different application contexts.

FAISS was quite accurate, efficient in dense indexing of small datasets using techniques like Flat Index and Hierarchical Navigable Small World. It always got outrun in lightweight memory use and query speed by Annoy, which offers competitive precision and scalability with very low memory usage. Among the various metrics employed with Annoy, it was found that Angular did really well in striking a nice balance between high precision and high query speed. The outcome of this study showed that indexing methods need to be adjusted to the characteristics of datasets and application requirements. While FAISS has the best results for precision-centric tasks, in general, Annoy is well-suited for memory- or computation-constrained environments because of its lightweight and scalable design. Also, the integration of deep learning feature extraction with ANN indexing underlines the potential of efficient retrieval techniques in large-scale data combined with modern neural architectures.

These findings set the base for further studies on ANN indexing methods for many domains. By appreciating the tradeoffs and relative strengths of FAISS and Annoy, researchers and practitioners are thus better positioned to deploy the best performing retrieval systems for their needs.